\documentclass{article} 

\usepackage{hyperref}
\usepackage{url}
\usepackage{times}  
\usepackage{helvet}  
\usepackage{courier}  
\usepackage{url}  
\usepackage{graphicx,wrapfig,subfigure}  
\usepackage{booktabs}

\usepackage{color}
\usepackage{amssymb}
\usepackage{amsmath}
\usepackage{graphicx}
\usepackage{amsfonts}
\usepackage{floatrow}

\usepackage[accepted]{icml2018}

\newfloatcommand{capbtabbox}{table}[][\FBwidth]

\DeclareMathOperator{\mean}{mean}
\DeclareMathOperator{\median}{median}

\icmltitlerunning{Unsupervised Document Embedding With CNNs}

\begin{document}

\twocolumn[
\icmltitle{Unsupervised Document Embedding With CNNs}

\icmlsetsymbol{equal}{*}

\begin{icmlauthorlist}
\icmlauthor{Chundi Liu}{equal,l6}
\icmlauthor{Shunan Zhao}{equal,l6}
\icmlauthor{Maksims Volkovs}{l6}
\end{icmlauthorlist}

\icmlaffiliation{l6}{layer6.ai}

\icmlcorrespondingauthor{Maksims Volkovs}{maks@layer6.ai}

\vskip 0.3in
]

\printAffiliationsAndNotice{\icmlEqualContribution}

\begin{abstract}

We propose a new model for unsupervised document 
embedding. Leading existing approaches either 
require complex inference or use recurrent neural 
networks (RNN) that are difficult to parallelize. 
We take a different route and develop a 
convolutional neural network (CNN) embedding model. 
Our CNN architecture is fully 
parallelizable resulting in over 10x speedup in 
inference time over RNN models. Parallelizable 
architecture enables to train deeper models
where each successive layer has increasingly larger
receptive field and models longer range semantic
structure within the document. We additionally propose 
a fully unsupervised learning algorithm to train this 
model based on stochastic forward prediction.
Empirical results on two public benchmarks show
that our approach produces comparable to state-of-the-art 
accuracy at a fraction of computational cost.

\end{abstract}

\section{Introduction}\label{sc:intro}

Document representation for machine reasoning is fundamental problem 
in natural language processing (NLP). A typical approach 
is to develop a document embedding model which produces fixed 
length vector representations that accurately preserve
semantic information within each document~\citep{blei2003lda,le2014distributed,
skipthought,arora2017embed,lin2017selftatt}. These models
can be supervised or unsupervised, and in this work we focus
on the unsupervised category where the models are trained 
using unlabeled text. The unsupervised approach is particularly 
attractive since large amount of unlabeled text is freely 
available on the Internet in virtually all major languages,
and can be used directly without expensive labeling or 
annotation. Moreover, since the embeddings can be utilized 
for a variety of tasks within the same NLP pipeline, even if 
labeling resources are available, it is difficult to determine 
what the target labels should be. Common tasks include sentiment 
and topic analysis, personalization and information retrieval, 
all of which would require different labels and embeddings if 
trained individually in a supervised fashion. Despite significant 
research effort in this area, the bag-of-words (BOW) and 
bag-of-ngrams approaches remain popular and still achieve 
highly competitive results~\cite{wang2012baselines}. However, 
BOW representations fail to capture similarities between words 
and phrases and as a result suffer from sparsity and 
dimensionality explosion. Moreover, by treating words as 
independent tokens, the temporal information is lost making 
it impossible to model long range semantic dependencies.

Recently, significant attention has been devoted to 
embedding approaches that use distributed representations 
of words~\cite{bengio2003neural,mikolov2013word2vec}.
Models within this category are trained to produce 
document embeddings from word representations, and either
jointly learn word representations during training or use a
pre-trained word model. The main advantage of these 
approaches is that they directly exploit
semantic similarities between words, and produce highly 
compact embeddings with state-of-the-art accuracy. 
Recent work has shown that embeddings with only 
several hundred dimensions achieve leading
accuracy on tasks such as topic/sentiment classification,
and information 
retrieval~\cite{le2014distributed,skipthought,lin2017selftatt}.

Within this category, popular approaches include weighted word
combination models~\cite{arora2017embed,chen2017embed}, 
doc2vec~\cite{le2014distributed} and recurrent neural 
network (RNN) models~\cite{skipthought,hill2016encoder,lin2017selftatt}.
The word combination models aim to directly aggregate word
representations in a given document through (weighted) 
averaging or another related function. Similarly to BOW,
these approaches are straightforward to implement and 
achieve highly competitive performance. Unlike BOW, the 
resulting embeddings are an order of magnitude smaller in
size and don't suffer from sparsity or dimensionality 
explosion problems. However, by averaging together word 
representations, temporal information is lost, and while
applying per word weights partially addresses this problem,
it doesn't eliminate it. One can easily come up with
examples of documents that contain nearly the same words, 
but have very different meaning due to word order. 
As such, averaging and other aggregation models 
that ignore word order are unlikely to perform well on 
the more complex NLP reasoning tasks.

doc2vec~\cite{le2014distributed} is another popular
unsupervised model which builds on the 
word2vec~\cite{mikolov2013word2vec} approach by incorporating 
document vectors that capture document specific semantic 
information. During training, both word and document vectors
are learned jointly, and word vectors are then held fixed
during inference. While accurate, this model 
requires iterative optimization to be conducted during inference. 
This involves computing multiple gradient updates and 
applying them to the document embedding with an 
optimizer of choice such as SGD. In high volume production 
environments, running such optimization for each new document 
can be prohibitively expensive. Moreover, as documents can vary 
significantly in length and word composition, it is difficult
to control for over/under-fitting without running further
diagnostics that add additional complexity.

Finally, RNN models~\cite{skipthought,hill2016encoder,lin2017selftatt}
address the inference problem by training a parametrized 
neural network model that only requires a deterministic 
forward pass to be conducted during inference. RNN embedding 
models ingest the document one word at a time and hidden 
activations (or their combination) after the entire document 
has been processed are then taken as the embedding.
This approach naturally addresses the variable length 
problem and provides a principled way to model temporal
aspects of the word sequence. However, the sequential nature of 
the RNN makes it difficult to leverage the full benefits of 
modern hardware such as GPUs that offer highly scalable parallel 
execution. This can significantly affect both training 
and inference speed. Consequently, most RNN embedding models are 
relatively shallow with only a few hidden layers. 
Moreover, many of the commonly used RNN achitectures,
such as LSTM~\citep{hochreiter1997lstm} and GRU~\citep{chung2014gru},
gate information from already seen input at each recurrence step. 
Repeated gating has an effect where more weight is placed 
on later words and the network can ``forget" earlier 
parts of the document~\citep{lai2015cnn}. This is not ideal
for language modeling where important information can occur
anywhere within the document.

In this work, we propose an unsupervised embedding model 
based on a convolutional neural network (CNN) that  
addresses the aforementioned problems. While CNNs have been 
utilized for supervised NLP tasks with considerable 
success~\cite{kim2014cnn,kalchbrenner2014cnn,dauphin2016language,conneau2017cnn},
little research has been done on the unsupervised problem. 
This work aims to address this gap. Specifically, we show that 
the convolutional architecture can be effectively utilized
to learn accurate document embeddings in a fully unsupervised
fashion. Convolutions are naturally parallelizable and do not
suffer from the memory problem of RNNs. This allows for 
significantly faster training and inference, leading to 
an order of magnitude inference speed-up over RNN 
embedding models. Faster training further 
enable us to explore deeper architectures that can model 
longer range semantic dependencies within the document.
We show that in all of these architectures the variable
length input problem can be effectively dealt with 
by using an aggregating layer between convolution and 
fully connected layers. This layer selects most salient 
information from the convolutional layers that is then 
combined in the fully connected layers to generate 
the embedding. Finally, we propose a new learning 
algorithm based on stochastic multiple word forward prediction.
This algorithm has few hyper parameters and is
straightforward to implement. In summary, our contributions 
are as follows:
\begin{itemize} 
\item We propose a CNN architecture for unsupervised
document embedding. This architecture is fully parallelizable
and can be applied to variable length input.

\item We develop a novel learning algorithm to train the CNN
model in a fully unsupervised fashion. The learning algorithm
is based on stochastic multiple word forward prediction, 
requires virtually no input pre-processing, and has few 
tunable parameters.

\item We conduct extensive empirical evaluation on public
benchmarks. Through this evaluation, we show that the 
embeddings generated by our CNN model produce comparable to
state-of-the-art accuracy at a fraction of computational cost.
\end{itemize}
\begin{figure*}[t]\centering
	\includegraphics[scale=0.63]{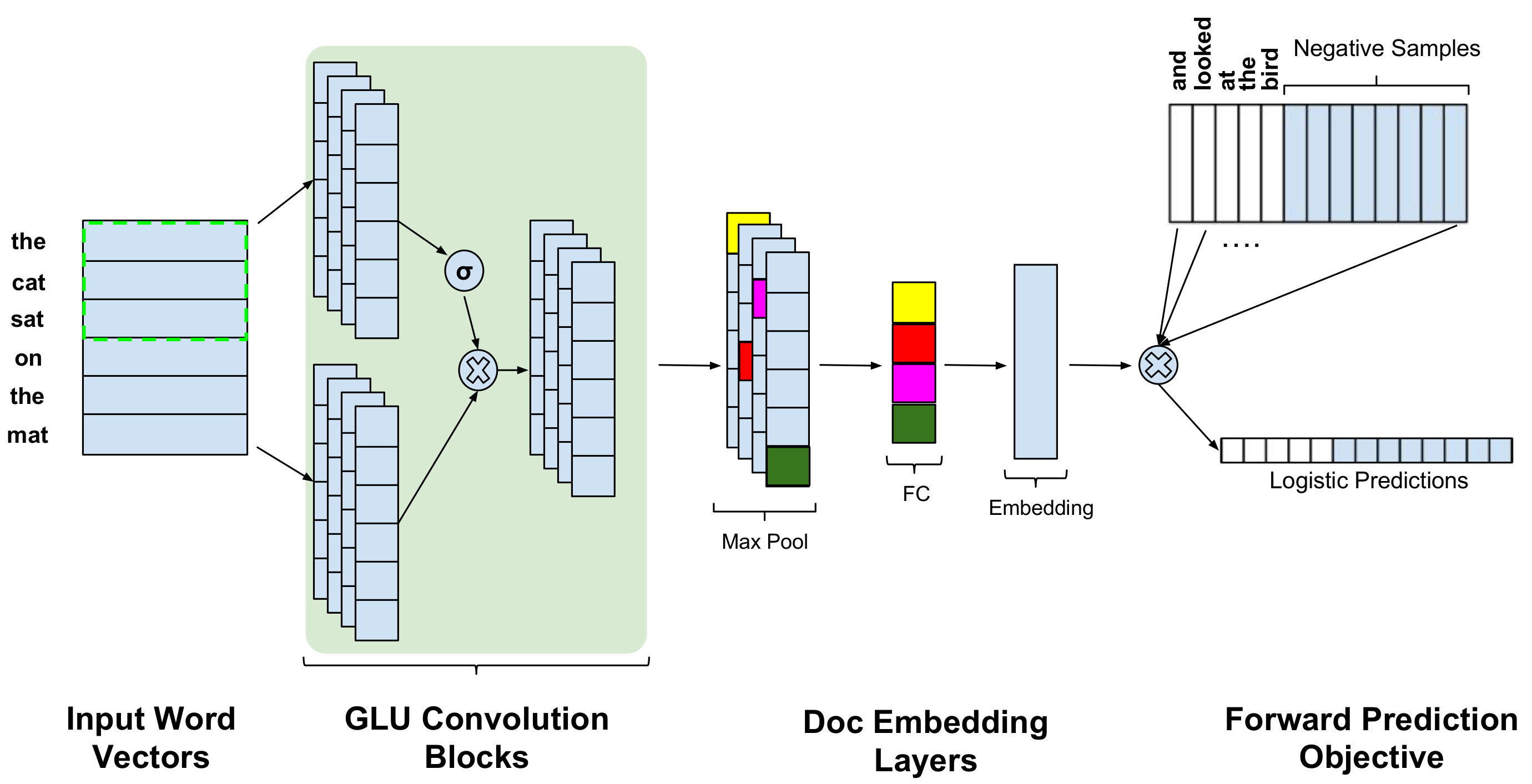}
	\vskip -0.2cm
	\caption{CNN embedding model diagram. The input 
	document is ``\texttt{the cat sat on the mat and looked at the bird}".
	During training, the document is partitioned at the word "\texttt{mat}''.
	Words up to "\texttt{mat}'' are used as input and passed
	through multiple layers of convolutions with 
	GLU activations~\cite{dauphin2016language}. The output of the 
	last GLU layer is passed through an aggregating function such as
	$\max$ pool or $\max_k$ pool. This operation converts variable length
	activation matrix into a fixed length one. Fixed length activations 
	are then passed through fully connected layers, the last of which outputs 
	the embedding. In this example, the model is trained to predict 
	the words after "\texttt{mat}'' so words in 
	"\texttt{and, looked, at, the, bird}'' are taken as positive 
	targets, and randomly sampled words are taken as negative targets.
	Dot products between the output embedding and target word 
	vectors are converted into probability using the logistic 
	function, and the model is updated with binary 
	cross-entropy objective.}
	\label{fig:model_diagram}
\end{figure*}

\section{Approach}

In this section we describe our model architecture, 
and outline learning and inference procedures. In a typical 
unsupervised document embedding problem, we are given a 
document corpus $\{D_1, \ldots, D_n\}$, where each 
document $D$ contains a sequence of words 
$D = w_1, \ldots, w_{|D|}$. The goal is to learn an embedding 
function $f:D \rightarrow \mathbb{R}^p$ that outputs a 
$p$-dimensional vector for every document. 
The embedding dimension is typically kept small, and highly
competitive performance has been demonstrated with 
$p < 1000$~\cite{le2014distributed}. Generated embeddings 
need to accurately summarize semantic/syntactic structure, 
and can be used as a primary document representation in 
subsequent NLP pipelines. Common use cases include efficient 
information storage and retrieval, and various supervised 
tasks such as sentiment analysis and topic classification. 
Note that besides documents in $\mathbb{D}$, we assume 
that no additional information is available and all 
training for $f$ is done in a fully unsupervised fashion.

As word sequences can't be used as input directly, a common 
approach is to first transform them into numeric format.
Recently, distributed 
representation~\cite{mikolov2013word2vec,pennington2014glove} has 
become increasingly more popular as it allows us to preserve temporal 
information, and doesn't suffer from sparsity or dimensionality 
explosion problems. Given a dictionary $\mathcal{V}$, 
each word $w \in \mathcal{V}$ is represented by a fixed
length vector $\phi(w) \in \mathbb{R}^m$. Concatenating 
together word vectors within the document provides the 
input representation:
\begin{equation}\label{eq:doc_rep}
\phi(D) = [\phi(w_1),...,\phi(w_{|D|})] 
\end{equation}
where $\phi(D)$ is an $m \times |D|$ input matrix.
Unlike bag-of-words, this representation fully preserves 
temporal order, is dense and captures semantic
similarity between words~\cite{mikolov2013word2vec}. 
Word vectors can either be learned together 
with the model or initialized using approaches such as
word2vec~\cite{mikolov2013word2vec} and 
glove~\cite{pennington2014glove}.
We use this format as our default input representation, and 
learn an embedding function $f(\phi(D), \theta)$ that maps 
$\phi(D)$ into a fixed length vector in $\mathbb{R}^p$,
where $\theta$ is a set of free parameters to be learned.

Given that $f$ operates on variable length input
word sequence, any subsequence of words within $D$ 
also forms a valid input. We use $D_{i:j} = w_i,w_{i+1}, \ldots, w_j$ 
to denote the subsequence from $i$'th to $j$'th 
word, and $\phi(D_{i:j})$ as the corresponding input 
representation:
\begin{equation}\label{eq:doc_subseq_rep}
\phi(D_{i:j}) = [\phi(w_i),\phi(w_{i+1}), \ldots, \phi(w_j)] 
\end{equation}
where $\phi(D_{i:j})$ in now an $m \times (j-i+1)$ matrix.
Passing the subsequence through $f$ yields the subsequence 
embedding $f(\phi(D_{i:j}), \theta)$. 
Intuitively, if $f$ has learned a ``good" representation 
function, then embedding $f(\phi(D_{i:j}), \theta)$ should 
accurately summarize semantic properties for any subsequence 
$D_{i:j}$. We extensively utilize this notion in our 
training procedure as it allows to significantly expand 
the training set by considering multiple subsequences 
within each document.

\subsection{Model Architecture}\label{sc:arch}

CNN models have recently been shown to perform well on supervised 
NLP tasks with distributed 
representations~\cite{kim2014cnn,kalchbrenner2014cnn,dauphin2016language,conneau2017cnn}, 
and are considerably more efficient than RNNs. Inspired by these 
results, we propose a CNN model for $f$.
Given an input matrix $\phi(D)$, we apply multiple layers of convolutions
to it. All convolutions are computed from left to right along the
word sequence, and the first layer is composed of $m \times d$ 
kernels that operate on $d$ words at a time. Stacking layers 
together allows each successive layer to model increasingly 
longer range dependencies with receptive fields that 
span larger sections of the input document. 
Similar to~\citet{dauphin2016language}, we found gated linear 
units (GLUs) to be useful for learning deeper models. Each
GLU convolutional layer is computed as follows:
\begin{equation}
h^l(x) = (h^{l-1}(x) * W^{l}  + b^{l}) \otimes 
\sigma(h^{l-1}(x) * V^{l}  + c^{l})
\end{equation}
where $h^l(x)$ is the output of $l$'th layer, $W^{l}$, $V^{l}$, 
$b^{l}$, $c^{l}$ are convolution weights, $\otimes$ is the 
element-wise product between matrices and $\sigma$ is the 
sigmoid function. Linear component of the GLU ensures that 
the gradient doesn't vanish through the layers, and 
sigmoid gating selectively chooses which information is 
passed to the next layer. This is analogous to the 
``forget gate" in the LSTM that selectively 
choses which information is passed to the next recurrence 
step. However, unlike LSTM, convolutions can be executed 
in parallel along the entire sequence, and GLU layers 
have no memory bias allowing the model to focus on any 
part of the input. Empirically, we found that learning 
with GLUs converged quicker and consistently produced 
better accuracy than ReLU activations. 

The output of the last GLU layer is an activation
matrix $h^l$ where each row corresponds to a convolutional kernel, 
and the number of columns varies with input length.
Fully connected layers can't be applied to variable length
activations so we investigate several ways to address this 
problem. The first approach is to apply zero padding to 
convert the input into fixed length:
\begin{equation}\label{eq:padding}
\phi(D) =
[\underbrace{\mathbf{0}, \ldots, 
\mathbf{0}}_{\max(k - |D|, 0)}, \phi(w_1), 
\phi(w_2), \ldots, \phi(w_{\min(|D|, k)})]
\end{equation}
where $\mathbf{0}$ is a zero vector and $k$ is the target length. 
Documents shorter than $k$ are left padded with zero vectors and 
those longer than $k$ are truncated after $k$ words. Analogous 
approach has been used by supervised CNN models for 
NLP~\citep{conneau2017cnn}. While conceptually simple and easy 
to implement, this approach has a drawback. For imbalanced 
datasets where document length varies significantly, it is 
difficult to select $k$. Small $k$ leads to information loss
for long documents while large $k$ results in wasted computation 
on short documents. 

To address this problem we note that the activation matrix
$h^l$ can be converted to fixed length by applying an
aggregating function along the rows. Common 
back-propagatable aggregating functions include $\max$, 
$\mean$ and $\median$. All of these convert arbitrary length
input into fixed length output. In this work we focus
on $\max$ which corresponds to the max pooling operation 
commonly used in deep learning models for computer vision
and other domains. The $\max$ operation is applied along
the rows of $h^l$ and produces an output vector with the same 
length as the number of convolutional kernels in layer $l$.
A further generalization of this procedure involves storing 
multiple values from each row of $h^l$ using an operator such as 
$\max_k$~\cite{kalchbrenner2014cnn} which outputs 
top-k values instead of top-1. The order in
which the $k$ maximum values occur is preserved 
in this operation, allowing the fully connected layers to 
capture additional temporal information~\cite{kalchbrenner2014cnn}. 

Another advantage of using the $\max$ operation is that by 
tracing the selected activations back through the network we 
can gain insight into which parts of the input word sequence 
the model is focusing on to generate the embedding. This can be used 
to interpret the model, and we show an example of this analysis in 
the experiments section. A popular alternative to max
pooling is attention layer~\citep{bahdanau2014attention}, 
and in particular self attention~\citep{lin2017selftatt}, 
where the rows of $h^l$ are first passed through a 
softmax functions and then self gated. However, 
this is beyond the scope of this paper and we leave it 
for future work.

Aggregating layer effectively deals with variable length input, 
and eliminates the need for document padding/truncation, 
which saves computation and makes the model more flexible.
After the aggregating layer, the fixed length activations 
are passed through fully connected layers, the last of which
outputs the $p$-dimensional document embedding. Full
architecture diagram with max pooling layer is shown 
in Figure~\ref{fig:model_diagram}.

\subsection{Learning and Inference}

We hypothesize that a ``good" embedding 
for a sequence of words $w_1, \ldots, w_i$ should be an
accurate predictor of the words $w_{i+1}, w_{i+2}, \ldots$ 
that follow. To accurately predict the next words the
model must be able to extract sufficient sufficient semantic 
information from the input sequence, and improvement in
prediction accuracy would indicate a better semantic 
representation. This forms the basis of our learning 
approach. We fix the embedding length $p$ to be the 
same as the word vector length $m$, and define the
probability that a given word $w$ occurs after the 
word sub-sequence $D_{i:j}$ using the sigmoid function:
\begin{equation}\label{eq:prob}
P(w | D_{i:j}, \theta) = 
\frac{1}{1 + e^{-\phi(w)^T f(\phi(D_{i:j}), \theta)}}
\end{equation}
where $\phi(w)^T f(\phi(D_{i:j}), \theta)$ is the dot 
product between the word vector $\phi(w)$ and the 
embedding for $D_{i:j}$. The probability is thus raised 
if word vector is similar to the embedding and lowered
otherwise. By optimizing both word vectors and embeddings,
the models learns a joint semantic space where dot product 
determines semantic relatedness.

During training, we optimize the model to perform well 
on this prediction task. The goal is to generate embeddings 
that can accurately predict the next words by having large 
dot product with each of them. Formally, given input 
document $D = w_1,...,w_{|D|}$ and prediction point 
$i$, we use subsequence $D_{1:i} = w_1,..., w_i$ to predict 
the next $h$ words $w_{i+1},..., w_{i+h}$. Framing this as 
a multi-instance binary classification problem, we optimize 
the cross entropy objective:
\begin{equation}\label{eq:loss}\small
\begin{aligned}
\mathcal{L}(i, D, \theta) = 
&-\sum^{i+h}_{\substack{j=i+1 \\ w_j \in D}} 
\log\left(P(w_j | D_{1:i}, \theta)\right)\\
&-\sum_{w \notin D} \log\left(1 - P(w | D_{1:i}, \theta)\right)\\
=&-\sum^{i+h}_{\substack{j=i+1 \\ w_j \in D}} 
\log\left(\frac{1}{1 + e^{-\phi(w_j)^T f(\phi(D_{1:i}), \theta)}}\right)\\
&-\sum_{w \notin D} \log\left(1 - 
\frac{1}{1 + e^{-\phi(w)^T f(\phi(D_{1:i}), \theta)}} \right)
\end{aligned}
\end{equation}
By minimizing $\mathcal{L}(i, D, \theta)$ we aim to raise 
the probability of the the $h$ words that follow $w_i$
and lower it for all other words. Expanding the forward 
prediction to multiple words makes the problem more challenging 
as it requires deeper understanding of the input sequence.
This in turn improves the embedding quality, and we found that 
predicting $h \geq 5$ words forward significantly 
improves accuracy.

In practice, it is expensive to compute the second term 
in Equation~\ref{eq:loss}. This term involves a summation 
over the entire word vocabulary for each document, which 
can be prohibitively large. We address this problem through sampling, 
and randomly sample a small subset of words to approximate this 
sum. Empirically, we found that using as few as $50$ word 
samples per document produced good results at a fraction of 
computational cost. Analogous approach is taken
with the prediction point $i$, instead of fixing it for 
each document we also use sampling. The 
sampling interval is $[\epsilon, |D|-h]$, and the 
prediction point is sampled uniformly within this interval. 
The lower bound $\epsilon > 1$ ensures that the model has 
sufficient context to do forward prediction. The upper bound 
$|D|-h$ ensures that there are at least $h$ words after $i$. 
In addition to simplicity, sampling prediction point has 
another advantage in that it forces the model to learn 
accurate embeddings for both short and long documents, 
acting as a regularizer and improving generalization.

The two sampling procedures lead to the learning algorithm
outlined in Algorithm~\ref{alg:learning}. This algorithm
is straightforward to implement, and only has three 
hyper parameters to tune. To accelerate 
learning we train with document mini-batches where 
the same prediction point and negative samples are used
for every document in the mini-batch, but are re-sampled 
across the mini-batches. Fixing the prediction 
point within the mini-batch allows to represent the input as 
a fixed length tensor which further improves parallelization. 
Moreover, to ensure that all training examples contribute equally
during optimization, we sample documents {\it without} 
replacement and count one epoch once all documents are
exhausted. An example of this learning procedure for a 
single document is shown in Figure~\ref{fig:model_diagram}. 
Here, input document is $D =$ \{\texttt{the cat sat on the mat 
and looked at the bird}\}, prediction point is $i = 6$ and
forward window is $h = 5$.
The subsequence $D_{1:6} =$ \{\texttt{the cat sat on the mat}\}
is passed through the CNN to generate the embedding 
$f(\phi(D_{1:6}), \theta)$. Dot products between this 
embedding and word vectors for target positive words 
\{\texttt{and, looked, at, the, bird}\} and randomly sampled
target negative words are converted into probabilities with
Equation~\ref{eq:prob}. The model is then updated using
the cross entropy objective from Equation~\ref{eq:loss}.
\begin{algorithm}[t] \caption{Learning Algorithm}
\begin{algorithmic}
\STATE {\bfseries Input:} $\{D_1,...,D_n\}$
\STATE {\bfseries Parameters:} offset $\epsilon$, 
forward prediction window $h$,
negative word sample size
\STATE {\bfseries Initialize:} CNN parameters $\theta$
\REPEAT[CNN optimization]
	\STATE sample document $D \in \{D_1,...,D_n\}$
	\STATE sample prediction point $i \in [\epsilon, |D|-h]$
	\STATE sample negative words $w \notin D$
	\STATE update $f$ using $\mathcal{L}(i, D, \theta)$ 
\UNTIL{convergence}
\STATE{\bfseries Output:} $f$
\label{alg:learning}
\end{algorithmic}
\end{algorithm}

Our proposed learning algorithm is different from existing
embedding models such as doc2vec and skip-thought. Unlike
these approaches, we use the {\it entire} subsequence $D_{1:i}$
to predict the words that follow $w_i$. This gives the model 
all the available information up to $w_i$, enabling it to 
capture richer semantic structure. In contrast, 
doc2vec uses a fixed context of 5-10 words and skip-thought is
trained on sentences. Training with larger context is 
possible in our model because CNN layers are fully 
parallelizable and allow for efficient forward and backward 
passes. Furthermore, instead of using fixed context or
tokenizing the input into sentences, we sample 
the prediction point. This eliminates the need for any 
input pre-processing while also keeping the context size dynamic.
We found this procedure to produce robust models even when 
the training data is relatively small. Finally, we only
train with forward word prediction while many existing models 
also do backward prediction. Backward prediction complicates 
training with feed-forward models, and goes against the 
natural language flow. We found that forward prediction 
training is sufficient to achieve highly competitive 
performance.

Once trained, inference in our model is done via a forward 
pass where the entire input document $D$ is passed through the CNN.
The resulting embedding $f(\phi(D), \theta)$ can then be used
in place of $D$ for further NLP tasks.
\begin{figure*}[t]
\centerline{
    \subfigure[]{
		\label{fig:num_layers}
    	\includegraphics[scale=0.23]{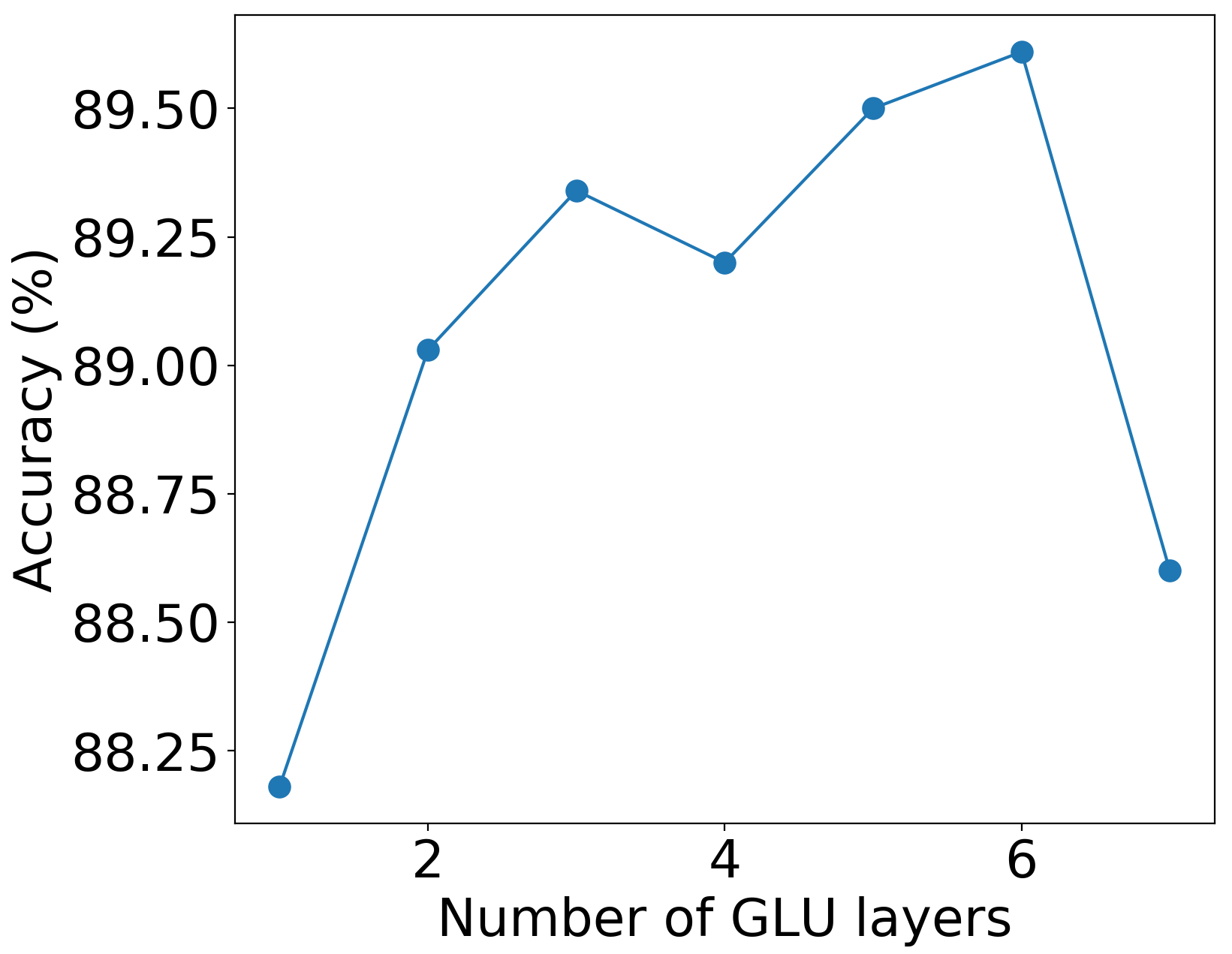}
	}
	\hskip 0.7cm
    \subfigure[]{
    	\label{fig:words_forward}
		\includegraphics[scale=0.23]{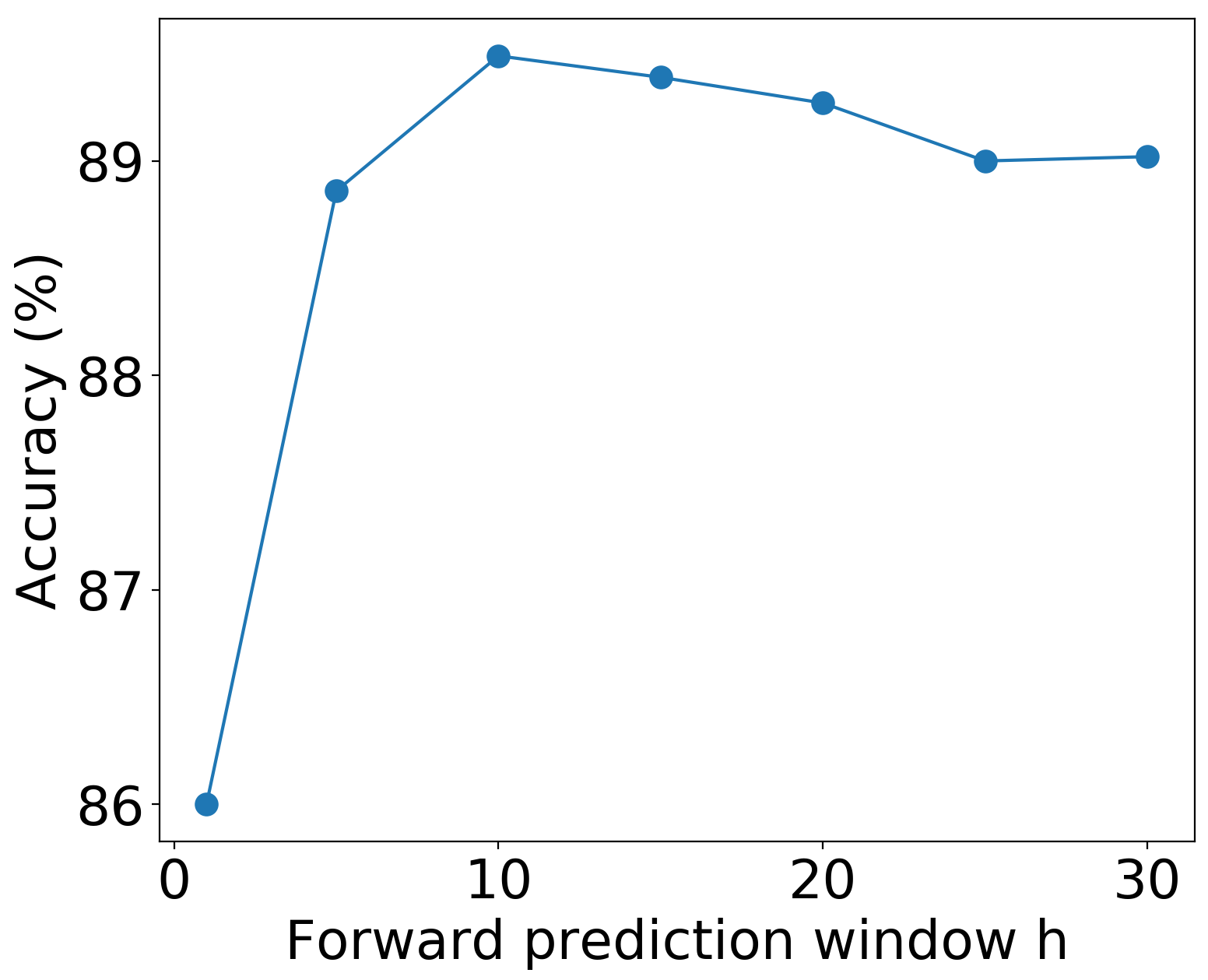}
    }
    \hskip 0.7cm
    \subfigure[]{
    	\label{fig:gating}
		\includegraphics[scale=0.23]{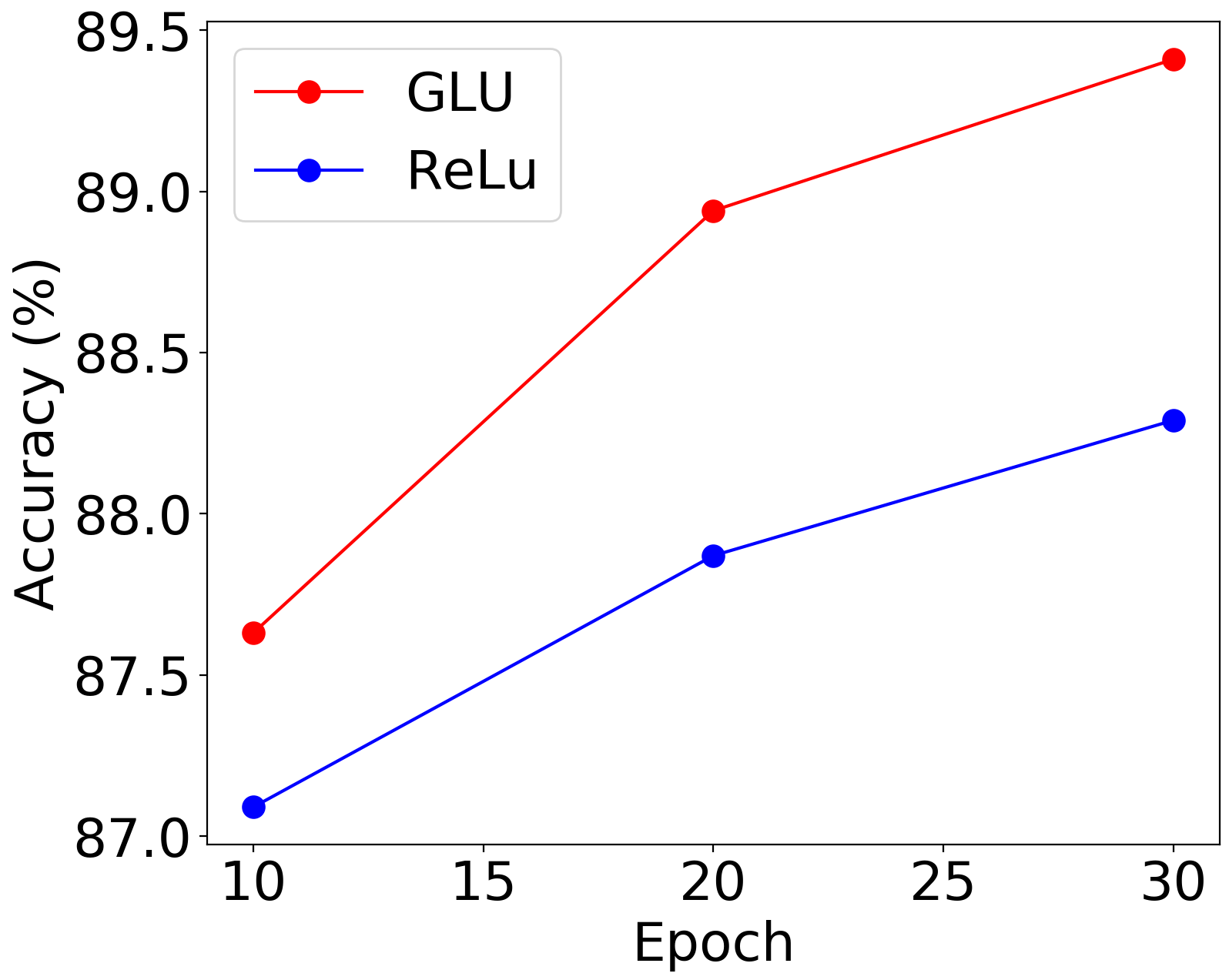}
    }
}
\vskip -0.7cm
\caption{IMDB performance analysis for various CNN architecture 
settings. Figure~\ref{fig:num_layers} shows accuracy vs number of
GLU layers. Figure~\ref{fig:num_layers} shows accuracy vs length
of the forward prediction window $h$. Figure~\ref{fig:gating} compares
GLU and ReLU layers.}
\label{fig:cnn_arch}
\end{figure*}

\section{Experiments}\label{sec:experiments}

To validate the proposed approach, we conducted extensive 
experiments on two publicly available datasets: IMDB~\citep{maas2011imdb}
and Amazon Fine Food Reviews~\citep{mcauley2013amazon}. We 
implemented our model using the TensorFlow library~\citep{abadi2016tensorflow}. 
All experiments were conducted on a server with 20-core Intel 
i7-6800K @ 3.40GHz CPU, Nvidia GeForce GTX 1080 Ti GPU, and 
64GB of RAM. We found that initializing word vectors with 
word2vec and then updating them during CNN training
resulted in faster learning and produced better performance. 
We use the pre-trained vectors taken from the
word2vec project page \footnote{\url{code.google.com/archive/p/word2vec}}, 
and thus fix the input word vector and output embedding 
dimensions to $m = p = 300$ for all models.
To address the variable length input problem, we experiment 
with both padding (CNN-pad) and max pooling (CNN-pool) 
approaches proposed in Section~\ref{sc:arch}. For max
pooling we additionally experiment with $\max_k$ pooling 
(CNN-pool-k) where $k$ largest values are retained for each
convolutional kernel. Through cross validation we found 
that setting $k = 3$ resulted in an optimal trade-off 
between accuracy and complexity, and use this value 
for all CNN-pool-k models.

Embeddings are evaluated by training a shallow classifier 
using the labeled training instances from each dataset, 
and we report the test set classification accuracy. 
The classifier has one hidden layer
with 100 hidden units and tanh activations. 
Classifier training is done using mini-batches of 
size 100 and SGD optimizer with momentum of 0.9. To 
make comparison fair we use the same classification 
set-up for all embedding models. Evaluation with 
labeled instances is analogous to previous work in this 
area~\cite{le2014distributed,skipthought}, 
and is aimed at validating whether the unsupervised 
models can capture sufficient semantic information 
to do supervised NLP tasks such as sentiment 
classification.

We compare our approach against the leading unsupervised
embedding models including doc2vec~\cite{le2014distributed},
doc2vecC~\cite{chen2017embed}, skip-thought~\cite{skipthought}
and SIF~\cite{arora2017embed}; all described in Section~\ref{sc:intro}. 
For each baseline we use the code from the respective authors, 
and extensively tune the model using parameter sweeps. 
Skip-thought code provides a pre-trained model (skip-thought-pr) 
that was optimized on a large book c1orpus, and we compare with 
this model as well as with skip-thought models tuned specifically 
for IMDB and Amazon datasets.

\subsection{IMDB}\label{sec:IMDB}

The IMDB dataset is one of the largest publicly available sentiment 
analysis datasets collected from the IMDB database. This dataset 
consists of 50,000 movie reviews that are split evenly into 25,000 
training and 25,000 test sets. There are no more than 30 reviews 
per movie to prevent the model from learning movie specific 
review patterns. The target sentiment labels are binarized: 
review scores $\leq 4$ are treated as negative and scores 
$\geq 7$ are treated as positive. In addition to labeled 
reviews, the dataset also contains 50,000 unlabeled 
reviews that can be used for unsupervised training.
Note that we remove test reviews during the unsupervised 
training phase, and train the embedding models using only 
the 50K unlabeled reviews + 25K training reviews. Removing the 
test reviews simulates the production environment where 
inference documents are typically not seen during the 
training phase. Our reported accuracy can thus differ 
from that reported by the previous work.

To understand the effects that the CNN architecture 
parameters have on performance we conduct extensive 
grid search, and record classification accuracy for 
various parameter settings. These experiments are
conducted with CNN-pad as they require many
training runs and CNN-pad is faster to train than
CNN-pool; results for CNN-pool exhibit similar patterns.
\setlength{\columnsep}{5pt}
\begin{wraptable}[7]{R}{0.24\textwidth}\small
	\vspace{-0.4cm}
    \begin{tabular}{lr}
    \toprule
    Model & tokens / s\\
    \midrule
    skip-thought-uni & 27,493\\
    skip-thought-bi & 14,374\\
    CNN-pad & {\bf 312,744}\\
    CNN-pool & 277,932\\
    \bottomrule
    \vspace{-0.9cm}
    \caption{Inference speed.}
    \label{tb:inf_speed}
    \end{tabular}
\end{wraptable}
Results for the most important parameters are 
shown in Figure~\ref{fig:cnn_arch}. Figure~\ref{fig:num_layers}
shows classification accuracy vs number of convolutional
GLU layers in the CNN. From this figure we see that
with the exception of four layers, the accuracy steadily
improves up to six layers. This indicates that depth
is useful for this task, and parallel execution in 
CNNs enables us to explore deeper architectures that
would not be possible with recurrent models.
We also found that after six layers the model 
would start to overfit, and aggressive 
regularization such as dropout and weight norm hurt accuracy. 
In the future work we aim to explore larger datasets as
well as additional regularization methods to address
this problem. 

Figure~\ref{fig:words_forward} shows accuracy vs
forward prediction window $h$. From the figure
it is seen that the accuracy significantly improves when
the model is trained to predict more than one word forward.
In particular, there is over 4\% gain when $h$ is increased from
$1$ to $10$, supporting the conclusion that the more difficult 
task of predicting multiple words leads to better embeddings. 
Finally, Figure~\ref{fig:gating} shows the difference in 
performance between GLU and ReLU layers. Across all epochs
GLUs consistently outperform ReLUs with relative improvement 
of 1\% to 2\%. 

Using these findings we select the following CNN architecture:
6 GLU layers with 900 kernels, batch 
normalization in each layer~\citep{ioffe2015batch} and residual 
connections every other layer~\citep{he2016resnet} to further 
accelerate learning. After GLU layers we apply max 
pooling (for CNN-pool) and a single fully connected layer 
that outputs the 300-dimensional embedding. This model
is trained to predict $h = 10$ words forward with 
$50$ negative word samples and prediction point offset 
$\epsilon = 10$ (see Algorithm~\ref{alg:learning}). 
Training is done using mini-batch gradient descent 
with batch size of $100$ and Adam optimizer~\cite{kingma2014adam}. 
Table~\ref{tb:inf_speed} shows inference speed in tokens 
(words) per second for this architecture as well as
doc2vec and uni/bi-directional skip-thought models. 
These results are generated by doing inference with batch size 1 to
remove the effects of batch CPU/GPU parallelization.
From the table it is our CNN architecture is over 10x 
faster than uni-directional skip-thought and over 20x 
faster than the bi-directional version. This is despite 
the fact that CNN has 7 hidden layers vs skip-thought RNN 
only has one. Similar results were reported 
by~\citep{dauphin2016language} on a related language 
modeling task, and clearly demonstrate the advantage 
of using the convolutional architecture.
\begin{table}
    \begin{tabular}{lc}
        \toprule
        Method & Accuracy\\
        \midrule
        Avg. word2vec~\cite{mikolov2013word2vec} & 86.25\\
        doc2vec~\cite{le2014distributed} & 88.22\\
        doc2vecC~\cite{chen2017embed} & 88.68\\
        skip-thought-pr~\cite{skipthought} & 82.57\\
        skip-thought~\cite{skipthought} & 83.44\\
        SIF~\cite{arora2017embed} & 86.18\\
        \midrule
        CNN-pad & 89.61\\ 
        CNN-pool & 90.02\\
        CNN-pool-k & {\bf 90.64}\\
        \bottomrule
    \vspace{-0.8cm}
    \caption{IMDB sentiment classification test set accuracy. 
    Test reviews are excluded during unsupervised model training.}
    \label{tb:imdb_results}
    \end{tabular}
\end{table}

Results for the sentiment classification experiments 
are shown in Table~\ref{tb:imdb_results}. From the table, 
we see that our best model CNN-pool-k outperforms all 
baselines and passes the difficult 90\% accuracy level. 
We also see that CNN-pool generally performs 
better than CNN-pad, suggesting that max pooling is more 
effective than padding and truncation. Both CNN-pool models
significantly outperform the RNN-based skip-thought 
approach. These results indicate that convolution with max 
pooling is a good alternative to recurrence for unsupervised 
learning with variable length textual input.

\subsection{Amazon Fine Food Reviews}\label{sec:AFFR}

The Amazon Fine Food Reviews dataset is a collection of 
568,454 reviews of Amazon food products left by users 
up to October 2012~\citep{mcauley2013amazon}. Each example 
contains full text of the review, a short summary, and a rating of 
1 to 5, which we use as the labels. This dataset does not 
come with a train-test split and is highly unbalanced. 
To address this, we perform our own split where after removing
duplicate reviews, we randomly sample 20,000 reviews from each 
class and from these, we randomly select 16,000 for training 
and 4,000 for testing. This produces training and test sets 
with 80,000 and 20,000 documents respectively.
\begin{table*}[t]\scriptsize\centering
    \begin{tabular}{ll}
        \toprule
        {\bf I found very little lobster in the can ...
        I also found I could purchase the same product at my 
        local Publix market at less cost.}\\
        0.755 \ \ This is a decent can of herring although not my favorite ... 
        I found the herring a little on the soft side but still enjoy them.\\
        0.738 \ \ You'll love this if you plan to add seafood yourself to 
        this pasta sauce ... 
        Don't use this as your only pasta sauce.  Too plain, boring ...\\
        0.733 \ \ I read about the over abundance of lobster in Maine ...
        I am not paying 3 times the amount of the lobster tails for shipping\\
        \midrule
        {\bf This drink is horrible ... The coconut water tastes like 
        some really watered down milk ... I would not recommend this to anyone.}\\
        0.842 \ \ I was really excited that coconut water came in flavors ... 
        but it is way too strong and it tastes terrible ... save your money ...\\
        0.816 \ \ wow, this stuff is bad.  i drink all the brands, all the time ... 
        It's awful ...  I'm throwing the whole case away, no way to drink this.\\
        0.810 \ \ This is the first coffee I tried when I got my Keurig.  
        I was so disappointed 
        in the flavor; tasted like plastic ... I would not recommend ...\\
        \midrule
        {\bf All I have to do is get the can out and my cat comes running.}\\
        0.833 \ \ Although expensive, these are really good.  My cat can't wait 
        to take his pills...\\
        0.787 \ \ My kitty can't get enough of 'em. She loves them so much that 
        she does anything 
        to get them ...\\
        0.770 \ \ Every time I open a can, my cat meows like CRAZY ... This is 
        the only kind of 
        food that I KNOW he likes. And it keeps him healthy.\\
        \bottomrule
    \end{tabular}
        \vspace{-0.3cm}
        \caption{Retrieval results on the Amazon dataset. 
        For each query review shown in bold we retrieve top-3 most similar 
        reviews using cosine distance between embeddings produced by our model.
        Cosine distance score is shown on the left for each retrieved result.}
        \label{tb:retrieval_results}
\end{table*}

We train our model using the same architecture and training method 
as in IMDB experiments, and compare against the same set of 
baseline approaches. All models are evaluated using binary
and 5-class classification tasks. For binary classification, 
documents with ratings 1 and 2 are treated as negative, 
4 and 5 as positive, and we discard documents labeled 3. 
However, all training documents, including those labeled 3, 
are used in the unsupervised training phase.
\begin{table}\centering
	\vspace{-0.2cm}
    \begin{tabular}{lcc}
        \toprule
        Method & 2-class & 5-class\\
        \midrule
        Avg. word2vec~\cite{mikolov2013word2vec} & 87.12 & 46.82\\
        doc2vec~\cite{le2014distributed} & 86.69 & 47.42\\
        doc2vecC~\cite{chen2017embed} & 89.25 & 51.17\\
        skip-thought-pr~\cite{skipthought} & {\bf 89.38} & 50.09\\
        skip-thought~\cite{skipthought} & 86.64 & 44.73\\
        SIF~\cite{arora2017embed} & 87.07 & 46.81\\
        \midrule
        CNN-pad & 87.06 & 50.42\\ 
        CNN-pool & 87.35 & 51.01\\  
        CNN-pool-k & 88.71 & {\bf 51.73}\\
        \bottomrule
        \vspace{-0.5cm}
        \caption{Amazon results for two and five class sentiment 
  			classification tasks.}
  	\label{tb:performance}
    \end{tabular}
\end{table}

Results for the classification task are shown in Table~\ref{tb:performance}.
From the table, we see that CNN-pool-k performs comparably to
the best baselines on both classification tasks. Max pooling 
again produces better performance than padding leading to 
conclusion that pooling should be used
as the default method to deal with variable length input.
Strong performance on the 5-class classification task
suggest that our model is capable of successfully learning 
fine-grained differences between sentiment directly from
unlabeled text. Overall, together with IMDB these results 
further indicate that the convolutional architecture 
is well suited for unsupervised NLP, and can be used 
to learn robust embedding models.

\subsection{Analysis}

A common application of document embedding is information 
retrieval~\citep{le2014distributed}, where the embedding vectors are 
indexed and used to quickly retrieve relevant results for a given query.
We use this approach to asses the quality of the embeddings that
our model generates. Using the Amazon dataset, we select several reviews
as queries and retrieve top-3 most similar results using embedding 
cosine distance as similarity measure. The results are shown in
Table~\ref{tb:retrieval_results}. From this table, we see that all
retrieved reviews are highly relevant to each query both in content
and in sentiment. The first group complains about seafood products, 
the second group is unhappy with a drink product, and the last group 
are cat owners that all like a particular cat food product. Interestingly, 
the product in the retried reviews varies, but both topic and sentiment stay
consistent. For instance, in the first group the three retrieved reviews
are about herring, seafood pasta and lobster. However, similar to the
query, they are all negative and about seafood. This indicates that
the model has learned the concepts of topic and sentiment without 
supervision, and is able to successfully encode them into embeddings.
\begin{figure}[t]\centering
	\vspace{-0.4cm}
	\includegraphics[scale=0.18]{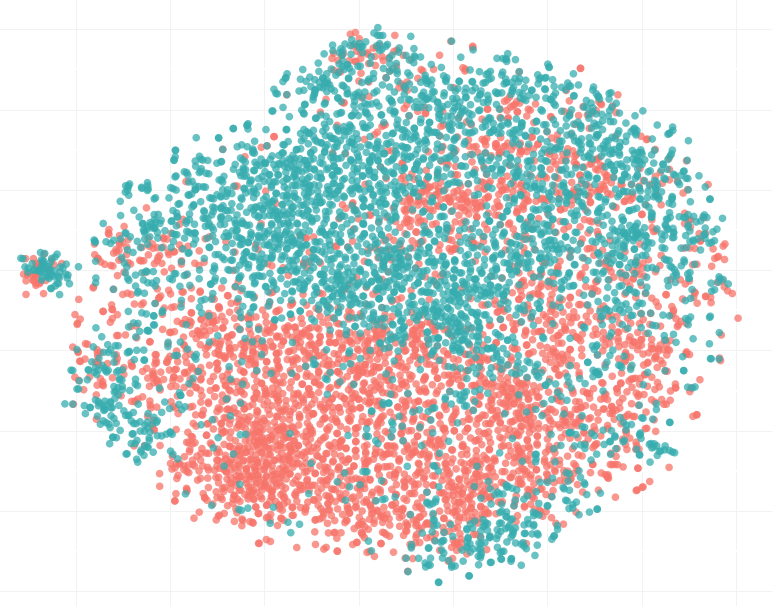}
	\vspace{-0.3cm}
	\caption{t-SNE representation of document embeddings 
	produced by our model for the IMDB test set. Points are 
	colored according to the positive/negative sentiment label.}
	\label{fig:imdb_tsne}
\end{figure}

To get further visibility into the embeddings, we applied 
t-SNE~\cite{maaten2008tsne} to the embeddings inferred for the 
IMDB test set. t-SNE compresses the embedding vectors into two 
dimensions and we plot the corresponding two dimensional points 
coloring them according to the sentiment label. This plot is shown in 
Figure~\ref{fig:imdb_tsne}. From the figure we see a distinct 
separation between sentiment classes where most negative reviews 
are near the top and positive reviews are 
at the bottom. This further validates that the model is able
to capture and encode sentiment information, making the two 
classes near linearly separable.

\section{Conclusion}

We presented a CNN model for unsupervised document embedding. 
In this approach, successive layers of convolutions are 
applied to distributed word representations 
to model longer range semantic structure within the document. 
We further proposed a learning algorithm based on stochastic
forward prediction. This learning procedure has few hyper 
parameters to tune and is straightforward to implement.
Our model is able to take full advantage of parallel
execution making it significantly faster than leading
RNN models. Experiments on public benchmarks show 
comparable to state-of-the-art accuracy at a fraction
of computational cost.


\bibliographystyle{icml2018}
\bibliography{icml2018_nlp_arxiv}

\end{document}